\documentclass[11pt]{article}

\usepackage[preprint]{acl}

\usepackage{times}
\usepackage{latexsym}

\usepackage[T1]{fontenc}

\usepackage[utf8]{inputenc}

\usepackage{microtype}

\usepackage{graphicx}

\usepackage[utf8]{inputenc} 
\usepackage[T1]{fontenc}    
\usepackage{hyperref}       
\usepackage{url}            
\usepackage{booktabs}       
\usepackage{amsfonts}       
\usepackage{nicefrac}       
\usepackage{microtype}      
\usepackage{colortbl}
\usepackage{array}

\usepackage{algorithm}
\usepackage{algorithmic}
\usepackage[utf8]{inputenc} 
\usepackage[T1]{fontenc}    
\usepackage{url}            
\usepackage{booktabs}       
\usepackage{amsfonts}       
\usepackage{nicefrac}       
\usepackage{microtype}      
\usepackage{bm}
\usepackage{enumitem}
\usepackage{graphicx}
\usepackage{amsmath}
\usepackage{subfigure}
\usepackage{amssymb}
\usepackage{caption}
\usepackage{longtable}
\usepackage{soul}
\usepackage{color}
\usepackage{appendix}
\usepackage{comment}

\usepackage{tikz}
\usetikzlibrary{trees}
\usetikzlibrary{positioning,fit}

\usepackage{forest}
\usepackage{array}

\title{A Survey of Automatic Prompt Optimization with Instruction-focused Heuristic-based Search Algorithm}

\author{Wendi Cui$^1$\thanks{The corresponding GitHub repository for this paper will be updated at \url{https://github.com/jxzhangjhu/Awesome-LLM-Prompt-Optimization}. Correspondence to wendi\_cui@intuit.com, jxzhangai@gmail.com.}, Zhuohang Li$^3$,  Hao Sun $^4$, Damien Lopez$^1$,  Kamalika Das$^{1,2}$\\ {\bf Bradley Malin$^{3,5}$, Sricharan Kumar$^{1,2}$, Jiaxin Zhang$^{1,2*}$}
\\ $^1$Intuit \quad  $^2$Intuit AI Research \quad $^3$Vanderbilt University \quad $^4$ University of Cambridge \\ \quad $^5$Vanderbilt University Medical Center\\
}

\begin{document}
\ifdefined\pdfoutput
  \maketitle
\fi

\begin{abstract}
Recent advances in Large Language Models (LLMs) have led to remarkable achievements, making \emph{prompt engineering} increasingly central to guiding model outputs. While manual methods (e.g., “chain-of-thought,” “step-by-step” prompts) can be effective, they typically rely on intuition and do not \emph{automatically} refine prompts. In contrast, \textbf{automatic prompt optimization} employing heuristic-based search algorithms can systematically explore and improve prompts with minimal human oversight. This survey proposes a \textbf{comprehensive taxonomy} of these methods, categorizing them by \emph{where} optimization occurs, \emph{what} is optimized,  \emph{what criteria} drive the optimization, \emph{which} operators generate new prompts, and \emph{which} iterative search algorithms are applied. We further highlight specialized \textbf{datasets and tools} that support and accelerate automated prompt refinement. We conclude by discussing key \textbf{open challenges} for future opportunities for more robust and versatile LLM applications. \end{abstract}

\section{Introduction}
\label{sec:introduction}
The rapid evolution of \textbf{Large Language Models} (LLMs) has catalyzed significant progress in diverse tasks \cite{bubeck2023sparks,yang2023harnessing}. As these models become more capable, the \emph{design of the prompt} used to interface with them has emerged as a crucial factor in terms of both prompt content and format~\cite{zhu2023promptbench}. Manual approaches such as \textit{“chain-of-thought”} \cite{wei2023cot} prompting or instructing the model to \textit{“Let's think step by step”} \cite{kojima2023stepbystep} can yield enhanced performance in certain scenarios, but these methods remain fundamentally reliant on human intuition and repeated trial-and-error.

In contrast, \textbf{automatic prompt optimization} aims to systematically discover and refine prompts, minimizing human efforts and potentially uncovering highly effective solutions that manual experimentation might overlook \cite{Zhou2023APE}. These techniques treat prompt design as a \emph{search problem}, wherein a \textbf{heuristic-based search algorithmic} process iteratively evaluates candidate prompts and adapts them based on performance feedback or other criteria \cite{pryzant2023APO, Chen2023InstuructZero, Yang2023OPRO}. Other lines of work applies reinforcement learning \cite{zhang2022tempera,deng2022rlprompt, sun2023query} or ensemble methods \cite{hou2023promptboosting,pitis2023boosted} to optimize prompts dynamically or adaptively. This survey focuses on heuristic-based search methods due to their ability to unify a broad range of strategies under an interpretable, modular framework that is compatible with both discrete and soft prompt spaces. We further concentrate on \textbf{instruction-focused} approaches, emphasizing the clarity and structure of the instruction while \textit{optionally} incorporating in-context examples. Instruction-based prompting continues to be the dominant paradigm, making it a practical and impactful focus. By narrowing our scope to this intersection, we aim to provide a coherent taxonomy and detailed analysis of methods that are both theoretically grounded and widely applicable.

We begin by examining the fundamental dimensions of \emph{where} optimization happens (optimization space) and \emph{what} is optimized (optimization target). We then review \emph{what criteria} to optimize (optimization criteria), recognizing that many practitioners now consider objectives beyond task performance. Then, we dive into how optimization happens by categorizing \emph{which operators} are used to create new prompt candidates and \emph{which iterative algorithm} guides the refinement loop. With a growing body of literature addressing these topics, we also review \textit{benchmarking datasets} covering a wide range of domains. Moreover, we survey a range of \textit{tools} that can automate or streamline prompt optimization workflows, enabling rapid iteration with less manual effort. We conclude by identifying open problems. Addressing these challenges will unlock reliable, adaptable, and ethical LLM applications.

\section{Preliminary}
\label{sec:heuristic_based_search}

\subsection{Prompt Composition}
\label{sec:composition_prompt}

Prompts generally contain two main components:

\begin{enumerate}[leftmargin=10pt, itemsep=1pt, topsep=1.5pt, partopsep=1.5pt]
    \item \textbf{Instruction:} The instruction is a human-readable statement describing the task or objective. An instruction establishes the \textit{intent} and \textit{context} for the model, guiding it toward the desired behavior for the particular task.
    \item \textbf{Examples:} In-context examples show the model how to map specific inputs to outputs. These examples clarify the nature of the task and can improve model performance \cite{Brown2020fewshot-icl}. 
\end{enumerate}

\noindent Prompts may include several examples or none. For this survey, we cover the \textbf{instruction-focused} automatic prompt optimization, excluding research on \textit{example-focused} optimization.

\subsection{Heuristic-based Search Algorithms}
\label{sec:heuristics_intro}

Heuristic-based search algorithms provide a practical framework for optimization problems such as automatic prompt optimization \cite{pryzant2023APO, Zhou2023APE, Fernando2023PromptBreeder}. Unlike brute-force methods that evaluate every possible variation, heuristic methods apply problem-specific knowledge or strategies to navigate the search space more efficiently \cite{blum2003metaheuristics, talbi2009metaheuristics}. Key steps in a heuristic-based approach typically involve:
\begin{enumerate}[leftmargin=10pt, itemsep=1pt, topsep=1.5pt, partopsep=1.5pt]
    \item \textbf{Initialization:} Generating one or more candidates (e.g., randomly, using partial domain knowledge, or by perturbing a known baseline).
    \item \textbf{Evaluation:} Measuring the performance of each candidate with respect to a chosen metric (e.g., accuracy on a validation set, or a scoring function provided by a user).
    \item \textbf{Selection and Update:} Applying operators or transitions to the current set of candidates to create new candidates with improved performance, diversity, or both.
    \item \textbf{Termination:} Determining when to stop (e.g., after a fixed number of iterations or once performance converges).
\end{enumerate}

\forestset{
    highlight1/.style={
        for tree={
            fill=green!8
        }
    },
    highlight2/.style={
        for tree={
            fill=orange!10
        }
    },
    highlight3/.style={
        for tree={
            fill=cyan!15
        }
    },
    highlight4/.style={
        for tree={
            fill=RoyalBlue!15
        }
    },
    highlight5/.style={
        for tree={
            fill=CornflowerBlue!8
        }
    },
    highlight6/.style={
        for tree={
            fill=RoyalBlue!15
        }
    },
    highlight7/.style={
        for tree={
            fill=CornflowerBlue!8
        }
    },
    highlight8/.style={
        for tree={
            fill=RoyalBlue!15
        }
    }
}

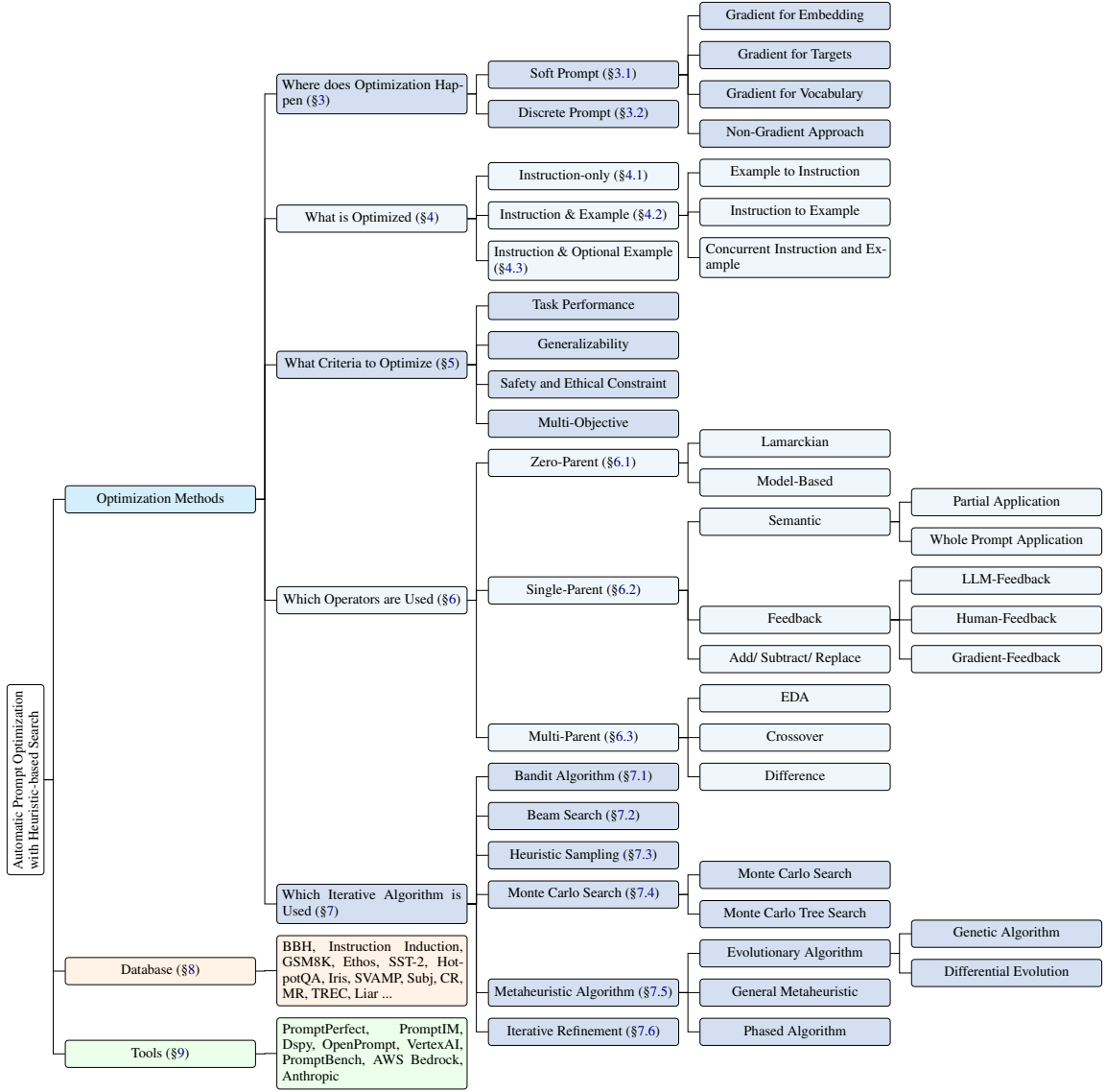
\begin{figure*}[ht]
    \centering
    \resizebox{0.95\textwidth}{!}{ 
        \begin{forest}
            for tree={
                grow=east,
                draw,
                text width=40mm,
                font=\small,
                edge path={
                    \noexpand\path [draw, \forestoption{edge}] (!u.parent anchor) -- +(2mm,0) |- (.child anchor)\forestoption{edge label};
                },
                parent anchor=east,
                child anchor=west,
                tier/.wrap pgfmath arg={tier #1}{level()},
                edge={ultra thin},
                rounded corners=2pt,
                align=center,
                text badly centered,
                sibling distance=20mm
            }
            [\parbox{40mm}{Automatic Prompt Optimization with Heuristic-based Search},
                transform shape,
                rotate=90,
                anchor=center,
                parent anchor=south,
                [Tools (§\ref{sec:tools_comparison}), highlight1,
                    [\parbox{40mm}{PromptPerfect, PromptIM, Dspy, OpenPrompt, VertexAI, PromptBench, AWS Bedrock, Anthropic}]
                ],
                [Database (§\ref{sec:common_datasets}), highlight2,
                    [\parbox{40mm}{BBH, Instruction Induction, GSM8K, Ethos, SST-2, HotpotQA, Iris, SVAMP, Subj, CR, MR, TREC, Liar ...}]
                ],
                [Optimization Methods, highlight3,
                    [\parbox{40mm}{Which Iterative Algorithm is Used (§\ref{sec:iteration_algorithms})}, highlight4,
                        [Iterative Refinement (§\ref{sec:iterative})],
                        [Metaheuristic Algorithm (§\ref{sec:metahurisitc}),
                            [Phased Algorithm],
                            [General Metaheuristic],
                            [Evolutionary Algorithm,
                                [Differential Evolution],
                                [Genetic Algorithm]
                            ]
                        ],
                        [Monte Carlo Search (§\ref{sec:montecarlo}),
                            [Monte Carlo Tree Search],
                            [Monte Carlo Search]
                        ],
                        [Heuristic Sampling (§\ref{sec:heuristicsampling})],
                        [Beam Search (§\ref{sec:beam})],
                        [Bandit Algorithm (§\ref{sec:bandit})]
                    ],
                    [Which Operators are Used (§\ref{sec:operators}), highlight5,
                        [Multi-Parent (§\ref{sec:multiparent}),
                            [Difference],
                            [Crossover],
                            [EDA]
                        ],
                        [Single-Parent (§\ref{sec:singleparent}),
                            [Add/ Subtract/ Replace],
                            [Feedback,
                                [Gradient-Feedback],
                                [Human-Feedback],
                                [LLM-Feedback]
                            ],
                            [Semantic,
                                [Whole Prompt Application],
                                [Partial Application]
                            ]
                        ],
                        [Zero-Parent (§\ref{sec:zeroparent}),
                            [Model-Based],
                            [Lamarckian]
                        ]
                    ],
                    [What Criteria to Optimize (§\ref{sec:criteria_optimize}), highlight6,
                        [Multi-Objective],
                        [Safety and Ethical Constraint],
                        [Generalizability],
                        [Task Performance]
                    ],
                    [What is Optimized (§\ref{sec:what_is_optimized}), highlight7,
                        [\parbox{40mm}{Instruction \& Optional Example (§\ref{sec:instruction_optional_example})}],
                        [Instruction \& Example (§\ref{sec:instructandexample}),
                            [\parbox{40mm}{Concurrent Instruction and Example}],
                            [Instruction to Example],
                            [Example to Instruction]
                        ],
                        [Instruction-only (§\ref{sec:instructonly})]
                    ],
                    [\parbox{40mm}{Where does Optimization Happen (§\ref{sec:where_optimization_happens})}, highlight8,
                        [Discrete Prompt (§\ref{sec:discrete})],
                        [Soft Prompt (§\ref{sec:softprompt}),
                            [Non-Gradient Approach],
                            [Gradient for Vocabulary],
                            [Gradient for Targets],
                            [Gradient for Embedding]
                        ]
                    ]
                ]
            ]
        \end{forest}
    }
    \caption{Taxonomy of Heuristic-based Search Algorithm in Automatic Prompt Optimization. Additional mapping of methods to taxonomy can be seen in Appendix Section \ref{sec:mapping}.}
    \label{fig:taxonomy}
\end{figure*}

\section{Where does Optimization Happen?}
\label{sec:where_optimization_happens}

Prompt optimization in LLMs can be broadly categorized into 	\textit{soft prompt space optimization} and \textit{discrete prompt space optimization}.

\subsection{Soft Prompt Space Optimization}
\label{sec:softprompt}

Soft prompt optimization operates in a continuous space, allowing for smooth adaptations with techniques such as \textit{gradient-based optimization}.  Different approaches vary in whether they utilize gradients and how they apply them.

\paragraph{Gradient for Embeddings} Using gradients for optimizing prompt embeddings is a widely used approach \cite{li2021prefix, zhang2021differentiable, sun2022black, sun2022bbtv2, sun2024retrieval}. \citet{concentrate2024li} uses gradient descent to optimize prompts leveraging a loss function tailored to improve prompts' generalization capability across diverse domains. 

To improve efficiency, some research chooses to estimate gradients, rather than directly computing them. ZOPO \cite{huw2024zopo} employs \textit{Zeroth-Order Optimization} to refine prompts without explicit gradient computation. It enables gradient estimation by using the Neural Tangent Kernel \cite{jacot2018ntk} in a derived Gaussian process, approximating optimization dynamics in neural networks.

\paragraph{Gradient-Based Target Selection}  Another line of research utilizes gradients to identify target tokens to replace within a prompt candidate \cite{rpo2024zhou, zou2023gcg, ps2024Zhao}. Greedy Coordinate Gradient (GCG) \cite{zou2023gcg} leverages gradient information to detect tokens that can minimize objective loss. It computes the gradient of the loss function with respect to the vector of each token, selecting the top-k tokens with the highest gradient values for modification.  \citet{ps2024Zhao} enhance GCG with \textit{Probe Sampling}, which accelerates prompt optimization by using two models: a smaller, faster "draft model" that assesses potential token replacement candidates and filters out unpromising ones; and a larger, more powerful "target model" which takes the filtered candidates for a full evaluation. Probe Sampling dynamically adjusts how many candidates are filtered by measuring the agreement between the draft and target models' rankings of a small "probe set" of candidates.  This adaptive filtering minimizes unnecessary computation for the large target model, leading to substantial speed gains.

\paragraph{Gradient for Vocabulary} In DPO \cite{dpo2024want}, gradients are estimated by a \textit{Shortcut Text Gradient} approach that continuously relaxes the categorical word choices to a learnable smooth distribution over the vocabulary using Gumbel Softmax trick. This allows for the computation of gradients through the non-differentiable embedding lookup table. By ultimately learning a distribution over the vocabulary, DPO iteratively improves the quality of the generated prompts.  

\paragraph{Non-Gradient Approach} Other works optimize in soft prompt space but do not employ a gradient-based approach. InstructZero \cite{Chen2023InstuructZero} employs \textit{Bayesian Optimization} to adjust prompt representations and propose new soft prompts without calculating gradients. 

\subsection{Discrete Prompt Space Optimization}
\label{sec:discrete}

Discrete prompt optimization treats prompts as fixed textual structures and refines them directly \cite{diao2022black, prasad2022grips}. Unlike soft prompt methods, which adjust prompts in a continuous space, discrete methods optimize prompts in a non-differentiable space.

While soft prompt methods leverage \textit{gradient}-based optimization, discrete approaches have developed \textit{gradient-like} strategies suited for non-differentiable settings. ProTeGi \cite{pryzant2023APO} employs an LLM-based feedback system to generate pseudo-gradients and utilizes beam search to iteratively refine prompts, effectively mimicking the refinement process in gradient-based methods.

Beyond \textit{gradient-like} approaches, other methods explore alternative optimization strategies. EvoPrompt \cite{Guo2023EVOPrompt} integrates evolutionary algorithms to iteratively refine prompts, employing semantic modification, crossover, and difference mechanisms for optimization. \citet{bpo2024cheng} trains a prompt-optimize model to rewrite prompts.

Overall, while soft prompt optimization excels in flexible and differentiable adjustments, discrete prompt optimization remains crucial for structured modifications where interpretability and explicit textual refinement are necessary.

\section{What is Optimized?}
\label{sec:what_is_optimized}

\subsection{Instruction-only Optimization}
\label{sec:instructonly}

Early approaches primarily focused on refining the instruction itself through techniques such as rephrasing, adding constraints, or incorporating additional context \cite{Yang2023OPRO, Hsieh2023AELP, pan2024plum}. \textit{After} instruction optimization, some approaches introduce examples randomly to enhance task performance \cite{Zhou2023APE, pryzant2023APO}. However, this choice often overlooks the interaction between added examples and instructions, resulting in suboptimal solutions \cite{wang2024mop, wang2024vertex}.

\subsection{Instruction \& Example Optimization}
\label{sec:instructandexample}

Recent research has increasingly focused on optimizing both instructions and examples. Existing approaches can be classified into three paradigms: 

\paragraph{Example to Instruction} This approach begins by selecting and preprocessing examples, which are then used to generate an appropriate instruction. MoP \cite{wang2024mop} adopts this strategy by clustering examples into \textit{Expert Subregions} and deriving specialized instructions for each cluster. This ensures that the instructions are well-aligned with the examples, improving task adaptation and overall effectiveness.

\paragraph{Instruction to Example}  
In this approach, an initial instruction is used to generate examples that align with the intended task. MIPRO \cite{mipro2024op} follows this methodology by employing a default instruction to create successful input-output pairs as examples. This ensures that the examples complement the instruction and reinforce the model’s understanding of the task.  

\paragraph{Concurrent Instruction and Example}  
This category focuses on optimizing both instructions and examples simultaneously. EASE \cite{wu2024ease} prioritizes selecting the best combination of instruction and examples from a pool of \textit{pre-defined candidates}, using bandit algorithms to identify the most effective prompt structure.  Whereas Adv-ICL \cite{advicl2024promptoptimizationadversarialincontext}, dynamically generates \textit{new instructions and examples}, refining both components iteratively with three models. By optimizing both elements in tandem, these methods ensure that instructions and examples are mutually reinforcing.  

\subsection{Instruction \& Optional Example}
\label{sec:instruction_optional_example}
Unlike prior approaches that strictly generate examples with instructions, PhaseEvo \cite{cui2024phaseevo} introduces a flexible framework capable of generating both few-shot and zero-shot prompts depending on what works best for the task. This adaptability allows the model to optimize performance dynamically, selecting whether examples are necessary based on empirical effectiveness.

\section{What Criteria to Optimize}
\label{sec:criteria_optimize}
In heuristic-based automatic prompt optimization, the \emph{objectives} or \emph{criteria} for refinement vary widely depending on the application domain. While early research predominantly focused on \emph{task performance}, growing interest in real-world deployments has led to broader optimization goals:

\begin{enumerate}[leftmargin=10pt, itemsep=1pt, topsep=1.5pt, partopsep=1.5pt]
    \item \textbf{Task Performance}  
    Most approaches prioritize task-specific metrics and optimize prompts to enhance performance on the given task. Some approaches use a \textit{validation set} to evaluate candidates \cite{Guo2023EVOPrompt}, while others use a \textit{surrogate model} of the objective function for candidate evaluation and selection \cite{mipro2024op, promptst2024chen}.

    \item \textbf{Generalizability}  
    Some methodologies extend beyond single-task performance, seeking prompts that generalize across multiple domains. \citet{concentrate2024li} introduce a \textit{Concentrate-focused framework} to improve the domain generalizability of prompts by leveraging internal information from deep model layers. Their findings indicate that \textit{prompts receiving higher attention from deep layers tend to generalize better} and that \textit{prompts with stable attention distributions enhance generalization}. Their approach optimizes for generalizability in both soft and discrete spaces.

    \item \textbf{Safety and Ethical Constraints}  
    Ensuring safety is a critical aspect of prompt optimization for large language models. Studies such as \textit{RPO} \cite{rpo2024zhou} emphasize the importance of designing prompt suffixes that resist adversarial manipulations and mitigate unintended behaviors. These safeguards are essential for defending against jailbreaking attempts.

    \item \textbf{Multi-Objective Optimization}  
    Multi-objective optimization plays a crucial role in balancing different priorities such as accuracy, efficiency, and safety. \textit{SOS} \cite{sinha2024sos} employs an \textit{interleaved} multi-objective evolutionary algorithms to optimize both task performance and safety where one objective is optimized first followed by another. In contrast, other approaches adopt a \textit{parallel} optimization strategy of all objectives, followed by Pareto Optimization to derive the most effective prompts across multiple objectives \cite{mopo2024resendiz, yang2023instoptima, baumann2024emo}.
\end{enumerate}

\section{Which Operators are Used}
\label{sec:operators}

For iterative optimization, generating new candidate prompts is essential. These generation methods, referred to as \textit{operators}, are categorized based on the number of \textit{parent prompts} needed. Parent prompts are existing prompts used to derive new candidates.

\subsection{Zero-Parent Operators}
\label{sec:zeroparent}

\paragraph{\textbf{Lamarckian}}  
The Lamarckian operator is an LLM-based operator that emulates the Lamarckian adaptation process, which refers to the idea of feeding back learned improvements (i.e., successful outputs or reasoning traces—phenotypes) into the prompt itself (genotype) to inform future generations. This mimics the Lamarckian notion of inheriting acquired traits \cite{Fernando2023PromptBreeder}. By analyzing concrete inputs that yield correct outputs, it attempts to reverse-engineer the instruction prompt. This operator is widely adopted in research, especially during initialization \cite{Zhou2023APE, huw2024zopo, wu2024ease, Fernando2023PromptBreeder, wang2024mop}.  MIPRO \cite{mipro2024op} extends this concept further by incorporating additional contextual information, such as identifying patterns within raw datasets, which enables LLMs to better comprehend task intentions. Table \ref{tab:lamarPrompt} presents an example of a Lamarckian operator.

\begin{table}[ht]
\centering
\small
\begin{tabular}{l}
\toprule 
\rowcolor{CornflowerBlue!10}I gave a friend an instruction and some input. \\
\rowcolor{CornflowerBlue!10}The friend read the instruction and wrote an \\
\rowcolor{CornflowerBlue!10}output for every one of the inputs.\\
\rowcolor{CornflowerBlue!10}Here are the input-output pairs:
\\ \rowcolor{CornflowerBlue!10}\\ 
\rowcolor{CornflowerBlue!10}\#\# Example \#\#  \\
\rowcolor{orange!8}\{\textit{input output pairs}\} 
\\\rowcolor{CornflowerBlue!10}
\\\rowcolor{CornflowerBlue!10}The instruction was:
\\
\bottomrule
\end{tabular}

\caption{Lamarckian Operator Prompt Example}
\label{tab:lamarPrompt}
\end{table}

\paragraph{\textbf{Model-Based}} 
This approach uses models to generate the next candidate. \citet{Chen2023InstuructZero} and \citet{sabbatella2023bayesian} build  probabilistic models of prompt performance using Bayesian Optimization. MIPRO \cite{mipro2024op} uses a Tree-structured Parzen estimator to build a surrogate model to select the instruction and example pair. INSTINCT \cite{lin2024useinstinct} uses a trained neural network for score prediction. Although such models do not take any parent prompts directly, they do tap into learning from previous prompts to enhance their performance.

\subsection{Single-Parent Operators}
\label{sec:singleparent}

\paragraph{\textbf{Semantic}} These operators generate candidates that share semantic meaning with their parent, either through the use of LLMs or alternative methods. Semantic operators can be categorized into:

\begin{itemize} [leftmargin=10pt, itemsep=1pt, topsep=1.5pt, partopsep=1.5pt]
    \item \textbf{Partial Application:} Semantic operators selectively modify specific sections of a prompt while maintaining the overall structure. This targeted approach enables precise adjustments, allowing for fine-tuning of particular aspects without altering the entire prompt. In AELP \cite{Hsieh2023AELP} and SCULPT \cite{kumar2024sculpt}, partial application of LLM-based semantic operators is utilized to systematically adjust key components of extensive prompts.

    \item \textbf{Whole Prompt Application:} Other semantic operators apply transformations to the entire prompt.  PhaseEvo \cite{cui2024phaseevo} uses an LLM-based semantic operator to perform the \textit{last-mile} optimization in a phased optimization process. FIPO \cite{lu2024fipo} finetunes a model to perform such rewriting. \citet{xu2022gps} translates a candidate to another language and back as a way to create new candidates. For soft prompts, \citet{shen2023free} uses a perturbation kernel to perturb the sampled candidate embeddings from the previous iteration to create new prompts.
\end{itemize}

\paragraph{\textbf{Feedback}} These methods utilize feedback from various sources to optimize prompts. They usually involve two steps: feedback generation and feedback application. Such feedback in a soft prompt space can be considered as the gradient. Based on the feedback generation process, such operators can be categorized as:

\begin{itemize} [leftmargin=10pt, itemsep=1pt, topsep=1.5pt, partopsep=1.5pt]
    \item \textbf{LLM-Feedback:}  These operators leverage LLMs to evaluate and refine prompts. Such an approach taps into LLMs'  self-reflection \cite{shinn2023selfreflect} ability to pinpoint deficiencies and suggest refinements, facilitating the automated creation of more robust and effective prompts through continuous self-improvement \cite{bpo2024cheng, dong-etal-2024-pace, ye2024pe2}.

    \item \textbf{Human-Feedback:}  Human feedback plays a crucial role in prompt optimization, providing nuanced, context-aware evaluations that automated systems may miss. PROMPST \cite{promptst2024chen} integrates human-designed rules to offer corrective feedback when errors arise, ensuring more precise refinements. APOHF \cite{lin2024apohf} leverages human preferences as an indicator for selecting the most effective prompts.
    
    \item \textbf{Gradient-Feedback:} Gradient-feedback involves using optimization techniques that adjust prompts based on gradient-based signals derived from the model's performance metrics. This approach is particularly effective for soft-prompt optimization and allows for precise and efficient adjustments \cite{dpo2024want, dln2023sordoni, rpo2024zhou}.
\end{itemize}

\paragraph{\textbf{Add/Subtract/Replace}}  These operators refine prompts by inserting, removing, or substituting elements \cite{prasad2022grips, juneja2024uniprompt, zhang2024sprig}. DPO \cite{dpo2024want} models each word as a genotype and applies these operators within an evolutionary algorithm framework for end-to-end prompt optimization.

\subsection{Multiple-Parent Operators}
\label{sec:multiparent}

\paragraph{\textbf{EDA}} These operators generate new candidates from multiple parent prompts. OPRO 
\cite{Yang2023OPRO} adds both parent prompts and their performance on the validation set as additional information. IPO \cite{ipo2024du} applies a similar strategy but on multimodal tasks. LCP \cite{li2024lcp} introduces contrastive examples to boost performance. Table \ref{tab:eda} is an example of an EDA operator.

\begin{table}[ht]
\centering
\small
\begin{tabular}{p{0.8\linewidth}}
\toprule
\rowcolor{CornflowerBlue!10}You are a mutator. Given a series of prompts, your task is to generate another prompt with the same semantic meaning and intentions.
\\
\rowcolor{CornflowerBlue!10}\\
\rowcolor{CornflowerBlue!10}\#\# Existing Prompts \#\#  \\
\rowcolor{orange!8}\{\textit{existing prompt}\} \\
\rowcolor{CornflowerBlue!10}\\

\rowcolor{CornflowerBlue!10}The newly mutated prompt is:\\

\bottomrule
\end{tabular}
\caption{EDA Prompt Example}
\label{tab:eda}
\end{table}

\paragraph{\textbf{Crossover}} This operator follows genetic algorithms and combines components from two-parent prompts to create new prompts \cite{baumann2024emo, yang2023instoptima, jin2024eot, Guo2023EVOPrompt}. The idea is to mix traits of both parents to compose a diverse candidate.

\paragraph{\textbf{Difference}} These LLM operators analyze differences between prompts to identify patterns for generating new candidates. EvoPrompt-DE \cite{Guo2023EVOPrompt} uses the Differential Evolution algorithm with such operators.

\section{Which Iterative Algorithm is Used}
\label{sec:iteration_algorithms}

Iterative algorithms are crucial in automatic prompt optimization. They guide the selection and application of operators to refine prompts effectively.

\subsection{Bandit Algorithms} 
\label{sec:bandit}

Bandit algorithms are a class of decision-making strategies designed to balance the exploration-exploitation trade-off. \citet{wu2024ease} formulates automatic prompt selection as a bandit problem, employing a neural bandit algorithm to predict the effectiveness of different sets of exemplars based on their embeddings. Similarly, \citet{bandit2024shi} introduces the BAI-FB framework, which efficiently explores and identifies the optimal prompt from a candidate pool while operating under a constrained budget. These approaches demonstrate the effectiveness of bandit-based methods in refining prompt selection and improving overall model performance.

\subsection{Beam Search}
\label{sec:beam}
The Beam Search method systematically expands a set of promising prompt candidates and prunes less effective ones, allowing efficient exploration of large search spaces. ERM \cite{yan2024erm}, ProTeGi \cite{pryzant2023APO} use beam search to iteratively select candidates for optimization. 

\subsection{Heuristic Sampling}
\label{sec:heuristicsampling}
Heuristic sampling is a method that utilizes rule-based strategies to efficiently select representatives from large sets of candidates minimizing computational resources while maintaining high accuracy. \citet{promptst2024chen}  employs heuristic-based sampling to prioritize the most promising prompts from an extensive pool, guided by human feedback to ensure their relevance and effectiveness.

\subsection{Monte Carlo Search} 
\label{sec:montecarlo}
\paragraph{Monte Carlo Search} Monte Carlo search is a probabilistic algorithm that uses random sampling to explore and evaluate possible actions or decisions within a given problem space, enabling the estimation of optimal strategies through repeated simulations. APE \cite{Zhou2023APE} leverages the Monte Carlo search to enhance prompt engineering by systematically exploring a vast array of potential prompts and assessing their effectiveness.

\paragraph{Monte Carlo Tree Seach} Monte Carlo Tree Search (MCTS) is a search algorithm that builds a search tree incrementally through a series of selection, expansion, simulation, and backpropagation steps. PromptAgent \cite{wang2023promptagent} constructs a search tree where each node represents a potential prompt. By keeping a state-action value function that calculates the potential rewards from following the path, the system iteratively refines prompts to enhance performance.

\subsection{Metaheuristic Algorithms}
Metaheuristic algorithms are high-level, problem-independent optimization strategies designed to efficiently explore large and complex search spaces where exact methods are infeasible. Inspired by natural processes such as evolution, physical systems, or social behavior, these algorithms guide the search toward optimal or near-optimal solutions through iterative improvement, balancing exploration and exploitation \cite{talbi2009metaheuristics}.

\label{sec:metahurisitc}
\paragraph{Evolutionary Algorithms}  
Evolutionary algorithms are widely utilized in prompt optimization, drawing inspiration from natural selection to iteratively refine prompts \cite{li2023spell, jin2024eot}. Two algorithms in this category are \textit{Genetic Algorithm (GA)} and \textit{Differential Evolution(DE)}: GA applies evolutionary principles, including mutation, selection, and crossover, to iteratively enhance prompts, ensuring gradual improvement across successive generations. DE generates new prompt candidates by utilizing the differences between existing solutions, promoting diversity while converging toward optimal solutions. EvoPrompt \cite{Guo2023EVOPrompt} systematically compares both across identical tasks, demonstrating that each algorithm excels in different scenarios.

\paragraph{General Metaheuristic} Additional metaheuristic algorithms such as Hill Climbing, Simulated Annealing, Tabu Search, Harmony Search, and others following metaheuristic principles are widely adopted for automatic prompt optimization as well \cite{ zhang2023peft, sun2023autohint, yang2024ampo, jin2024apeer, lin2024useinstinct, gao2025maps, tang2025gpo}. \citet{pan2024plum} specifically conducted a comparison among them.

\paragraph{Phased Algorithms} \citet{cui2024phaseevo} proposes a phased algorithm adopting a metaheuristic framework to increase the efficiency of the optimization process. By creating four phases balancing exploration and exploitation, they achieve several magnitudes of efficiency improvements compared to other algorithms in terms of LLM calls.

\subsection{Iterative Refinement}
\label{sec:iterative}
Iterative Refinement refers to the other algorithms that repeatedly use different operators to refine prompts. Gradient descent is a widely-adopted example \cite{huw2024zopo, dpo2024want, rpo2024zhou}.

\section{Common Datasets Used}
\label{sec:common_datasets}

Considering the broad applicability of prompt optimization, a variety of databases across different domains were used, as shown in Table \ref{tab:dataset_categorization} in the Appendix. The two most common ones are:

\begin{itemize}[leftmargin=10pt, itemsep=1pt, topsep=1.5pt, partopsep=1.5pt]
    \item \textbf{BBH (Big-Bench Hard)}~\cite{srivastava2023beyond}: 
    A subset of the broader Big-Bench project, BBH is designed to probe the limits of language models with especially challenging or nuanced tasks. 

    \item \textbf{Instruction Induction}~\cite{Honovich2022Instuct}: 
    This dataset explicitly focuses on inferring new task instructions from examples, making it particularly relevant for evaluating \emph{instruction-based} prompt optimization approaches.
\end{itemize}

\section{Common Tools}
\label{sec:tools_comparison}

Automatic prompt optimization tools aim to accelerate and streamline the optimization process. 
Below, we provide an overview of notable tools and their key characteristics. Table \ref{tab:prompt-tools} gives a high-level overview of these tools.

\begin{table*}[ht]
    \centering
    \small
    \begin{tabular}{l|c|c|c}
        \toprule
        \textbf{Tool} & \textbf{Optimization Space} & \textbf{Key Features} & \textbf{Open Source} \\
        \midrule
        PromptPerfect & Discrete & Web-based, optimized for user queries & No \\
        \midrule
        PromptIM & Discrete &  Iterative refine with human in the loop & Yes \\
        \midrule
        Dspy \cite{khattab2023dspy} & Discrete & Task decomposition and example bootstrap & Yes \\
        \midrule
        OpenPrompt \cite{ding2021openprompt} & Soft/Discrete & Predefined templates for prompt learning & Yes \\
        \midrule
        Vertex AI \cite{wang2024vertex} & Discrete & Google Cloud-based optimization & No \\
        \midrule
        PromptBench \cite{zhu2023promptbench} & 
        Discrete & Test prompt robustness & Yes \\
        \midrule
        AWS Bedrock & Discrete & Playground with evaluation and A/B testing & No \\
        \midrule
        Anthropic Claude & Discrete & Interactive editor with live feedback & No \\
        \bottomrule
    \end{tabular}
    \caption{Comparison of automatic prompt optimization tools.}
    \label{tab:prompt-tools}
\end{table*}

\paragraph{PromptPerfect} PromptPerfect is a commercial platform that offers automated prompt optimization services. It allows users to input a prompt and target a specific LLM. The platform then uses its internal algorithms to refine and improve the prompt. It provides a user-friendly interface and is designed to be accessible even to users without deep technical expertise.

\paragraph{PromptIM} PromptIM is an experimental open-source library.  Given an initial prompt, a dataset, and evaluators, PromptIM iteratively refines the prompt using a meta prompt to suggest improvements based on evaluation scores.  It integrates with LangSmith for data and prompt management and supports optional human feedback. Contrary to other tools, PromptIM prioritizes keeping humans in the loop throughout the optimization process.

\paragraph{DSPy} DSPy is a framework developed by Stanford researchers that takes a more declarative approach to building complex LLM applications.  Instead of directly writing prompts, users define the desired "program" as a series of declarative steps. DSPy implements MIPRO \cite{mipro2024op} and uses LLMs to generate and optimize the underlying prompts to fulfill the program’s goals. This approach allows for more structured and modular development of LLM applications and facilitates prompt optimization as part of the program execution. DSPy emphasizes the decomposition of complex tasks into simpler sub-tasks and is widely used in research \cite{together2024soylu}.

\paragraph{OpenPrompt} OpenPrompt \cite{ding2021openprompt} is an open-source toolkit specifically designed for prompt-learning. It provides a standardized framework for implementing and experimenting with various templating, verbalizing, and optimization strategies. OpenPrompt offers pre-defined templates for different prompting methods, such as prefix-tuning and P-tuning.  Its combinability across different Pretrained LMs, task formats, and optimization methods makes it a valuable tool.

\paragraph{Vertex AI Prompt Optimizer} Google Cloud's Vertex AI platform offers a Prompt Optimizer as part of its suite of tools. This service allows users to experiment with and optimize prompts following learning from \citet{wang2024vertex}. Integrated within the Vertex AI ecosystem, the Prompt Optimizer benefits from Google's cloud infrastructure and provides a scalable solution for prompt optimization tasks.

\paragraph{PromptBench} 
PromptBench \cite{zhu2023promptbench} is an open-source benchmark designed to evaluate the robustness of prompts under adversarial perturbations. Rather than introducing new task datasets, PromptBench applies systematic transformations—such as instruction negation or paraphrasing—to existing prompts across a variety of standard NLP datasets. It helps researchers assess how well prompt optimization methods preserve model performance under distributional shift.

\paragraph{AWS Bedrock} 
AWS Bedrock includes a Prompt Engineering Playground within its cloud platform, allowing users to prototype and evaluate prompts across multiple foundation models. The interface supports A/B testing, real-time inference, and evaluation metrics. While not open-source, it provides a practical environment for optimizing and comparing prompt variants in production-ready workflows.

\paragraph{Anthropic Claude Prompt Tools} 
Anthropic's Claude platform offers interactive tools for prompt optimization directly through its web interface. These tools provide live feedback, suggest rewrites, and support prompt experimentation tailored specifically to Claude models. While proprietary, they are useful for developers seeking to iteratively refine instructions with guidance grounded in Claude's internal safety and helpfulness principles.

\section {Open Challenges}
\label{sec:challenge}

\paragraph{Soft to Discrete Projection}  Soft prompt spaces enable continuous optimization but often lack interoperability provided by discrete prompts. To address this, some methods project soft embeddings back into discrete space. \citet{huw2024zopo}, \citet{wen2023hardpromptseasygradientbased} and \citet{ps2024Zhao} adopt a pre-generated finite set of unique candidates, where gradient-updated embeddings are mapped back to the closest entry in this set. However, this approach heavily depends on the quality of the pre-generated set.  Another approach leverages an open-source LLM as a converter to translate soft prompts into discrete instructions \cite{Chen2023InstuructZero, lin2024useinstinct}, offering a more flexible and adaptive solution. Further research in this area could enhance both optimization effectiveness and interpretability.

\paragraph{Dynamic N-shot Selection}  While \textit{Instruction \& Example} paradigms have shown significant improvements by jointly optimizing examples and instructions \cite{wang2024vertex, mopo2024resendiz}, recent findings indicate that few-shot prompting does not always enhance performance and can \textit{"consistently degrades its performance"} \cite{deepseek2025r1}. This highlights the necessity of \textit{Instruction \& Optional Example} paradigm which dynamically selects between few-shot and zero-shot prompting based on empirical effectiveness rather than a fixed strategy. Initial steps in this direction have been explored by \citet{cui2024phaseevo}, and future optimization approaches should emphasize adaptability, tailoring prompt structures to specific tasks for optimal performance.

\paragraph{Concurrent Optimization}  For complex tasks using LLMs, multiple agents might be involved \cite{sac32024zhang, cui2024dcr, ski2024zhang}. Traditionally, humans will define the scope of each agent and optimize them separately. Recent research has introduced automated concurrent optimization, which optimizes multiple prompts concurrently.  DLN-2 \cite{dln2023sordoni} allows concurrent optimization of two prompts by considering both LLMs as probabilistic layers in a network. It treats the first prompt’s output as a latent variable. Using variational inference to approximate the latent variable with a simpler distribution, DLN-2 optimizes the Evidence Lower Bound to refine both prompts. MIPRO \cite{mipro2024op} extends this concept to multi-stage optimization, treating each stage as a module and using Bayesian Search to identify the best prompt combinations. These methods represent a shift towards concurrent prompt optimization, reducing human effort while improving adaptability for complex task scenarios.

\paragraph{Additional Challenges} 
Beyond the challenges discussed earlier, several open issues remain critical. \textit{Multi-objective optimization} continues to be a complex area, requiring methods that can balance performance, safety, generalizability, and efficiency simultaneously \cite{sinha2024sos, mopo2024resendiz, yang2023instoptima}. Recent work explores Pareto-front approximations, but reliable aggregation of heterogeneous metrics remains unsolved. In addition, \textit{Scalability across Domains and Tasks} is limited by overfitting to specialized datasets. General-purpose optimizers must learn transferable representations or search strategies applicable in diverse settings. Another area is \textit{On-line fashion optimization}. Existing methods take thousands of API calls \cite{cui2024phaseevo} or require specific training \cite{huw2024zopo}, making it impractical for online optimization. Incremental update rules and memory-efficient surrogate models could empower near real-time regimes.

\section{Conclusion}
\label{sec:conclusion}

This survey has explored heuristic-based search algorithms for \textit{Automatic Instruction-Focused Prompt Optimization}, organizing methods along five key dimensions: the optimization space, the optimization target, the optimization criteria, the operators, and the iterative algorithms. The goal was to allow \textbf{mixing and matching} components like a toolkit, enabling the design of new prompt optimization pipelines by combining different operators with various search or learning algorithms. To make this more concrete, one might think of our framework like a recipe book: operators are the ingredients (e.g., "add", "replace", "rephrase"), and the iterative algorithms are the cooking methods (e.g., "bake" with genetic algorithms, "slow simmer" with beam search). Different combinations yield different flavors—and innovations. We hope this survey could jump-start researchers in understanding the existing landscape and inspire new research on the practical application of LLMs.

\section{Limitations}
The work does not cover In Context Learning optimization or methods using reinforcement learning. We also focus on works after 2023 so previous work is not fully covered.  Such space can be expanded for future works.

\section*{Acknowledgments}
This work includes contributions from Vanderbilt University researchers, supported by funding from Intuit.

\bibliography{custom}

\clearpage
\appendix

\onecolumn

\section{Methods Categorization based on Taxonomy}
\label{sec:mapping}
Below are the categorizations for methods surveyed in this paper based on the taxonomy.

\vspace{10mm}
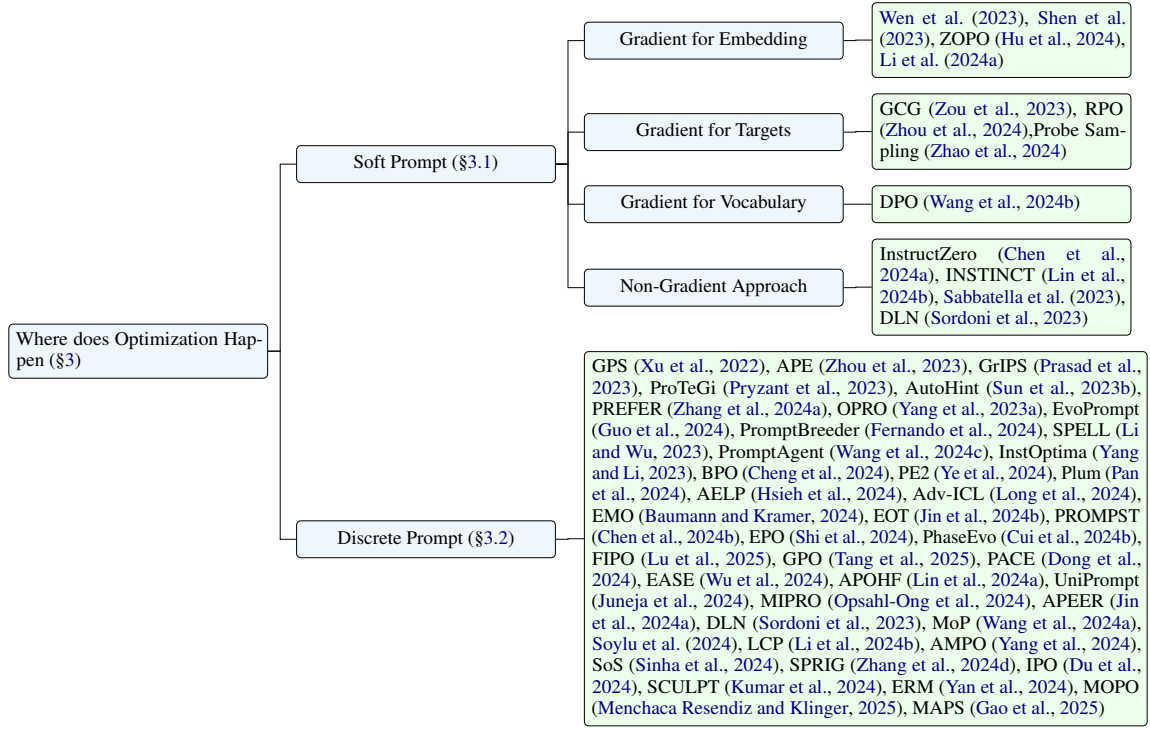
\begin{figure}[ht]
    \centering
    \resizebox{0.95\textwidth}{!}{ 
        \begin{forest}
            for tree={
                grow=east,
                draw,
                text width = 40mm,
                font=\small,
                edge path={
                    \noexpand\path [draw, \forestoption{edge}] (!u.parent anchor) -- +(2mm,0) |- (.child anchor)\forestoption{edge label};
                },
                parent anchor=east,
                child anchor=west,
                tier/.wrap pgfmath arg={tier #1}{level()},
                edge={ultra thin},
                rounded corners=2pt,
                align=center,
                text badly centered, 
                sibling distance=20mm,
            }
            [\parbox{40mm}{Where does Optimization Happen (§\ref{sec:where_optimization_happens})}, highlight7,
                [Discrete Prompt (§\ref{sec:discrete}),
                    [\parbox{90mm}{
                        GPS \cite{xu2022gps}, APE \cite{Zhou2023APE},
                        GrIPS \cite{prasad2022grips}, 
                        ProTeGi \cite{pryzant2023APO}, AutoHint \cite{sun2023autohint}, PREFER \cite{zhang2023peft}, OPRO \cite{Yang2023OPRO}, EvoPrompt \cite{Guo2023EVOPrompt}, PromptBreeder \cite{Fernando2023PromptBreeder}, SPELL \cite{li2023spell}, PromptAgent 
                        \cite{wang2023promptagent}, 
                        InstOptima \cite{yang2023instoptima}, BPO \cite{bpo2024cheng}, PE2 \cite{ye2024pe2}, Plum \cite{pan2024plum}, AELP \cite{Hsieh2023AELP}, Adv-ICL \cite{advicl2024promptoptimizationadversarialincontext}, EMO \cite{baumann2024emo}, EOT \cite{jin2024eot}, PROMPST \cite{promptst2024chen}, EPO \cite{bandit2024shi}, PhaseEvo \cite{cui2024phaseevo}, FIPO \cite{lu2024fipo}, GPO \cite{tang2025gpo}, PACE \cite{dong-etal-2024-pace}, EASE \cite{wu2024ease}, APOHF \cite{lin2024apohf}, UniPrompt \cite{juneja2024uniprompt}, MIPRO \cite{mipro2024op}, APEER \cite{jin2024apeer}, DLN \cite{dln2023sordoni}, MoP \cite{wang2024mop}, \citet{together2024soylu}, LCP \cite{li2024lcp}, AMPO \cite{yang2024ampo}, SoS \cite{sinha2024sos}, SPRIG \cite{zhang2024sprig}, IPO \cite{ipo2024du}, SCULPT \cite{kumar2024sculpt}, ERM \cite{yan2024erm}, MOPO \cite{mopo2024resendiz}, MAPS \cite{gao2025maps}
                        }, text width=90mm, tier=odd, highlight1
                    ]
                ]
                [Soft Prompt (§\ref{sec:softprompt}),
                    [Non-Gradient Approach,
                        [\parbox{40mm}{
                        InstructZero \cite{Chen2023InstuructZero}, INSTINCT \cite{lin2024useinstinct}, \citet{sabbatella2023bayesian}, DLN \cite{dln2023sordoni}
                        }, highlight1]
                    ]
                    [Gradient for Vocabulary,
                        [\parbox{40mm}{DPO \cite{dpo2024want}}, highlight1
                        ]
                    ]
                    [Gradient for Targets,
                        [\parbox{40mm}{GCG \cite{zou2023gcg}, RPO \cite{rpo2024zhou},Probe Sampling \cite{ps2024Zhao}
                        }, highlight1]
                    ]
                    [Gradient for Embedding,
                        [\parbox{40mm}{\citet{wen2023hardpromptseasygradientbased},  \citet{shen2023free}, ZOPO \cite{huw2024zopo}, \citet{concentrate2024li}
                        }, highlight1]
                    ]
                ]
            ]
        \end{forest}
    }
    \caption{\textit{"Where does optimization happen"} Categorization }
    \label{fig:space}
\end{figure}

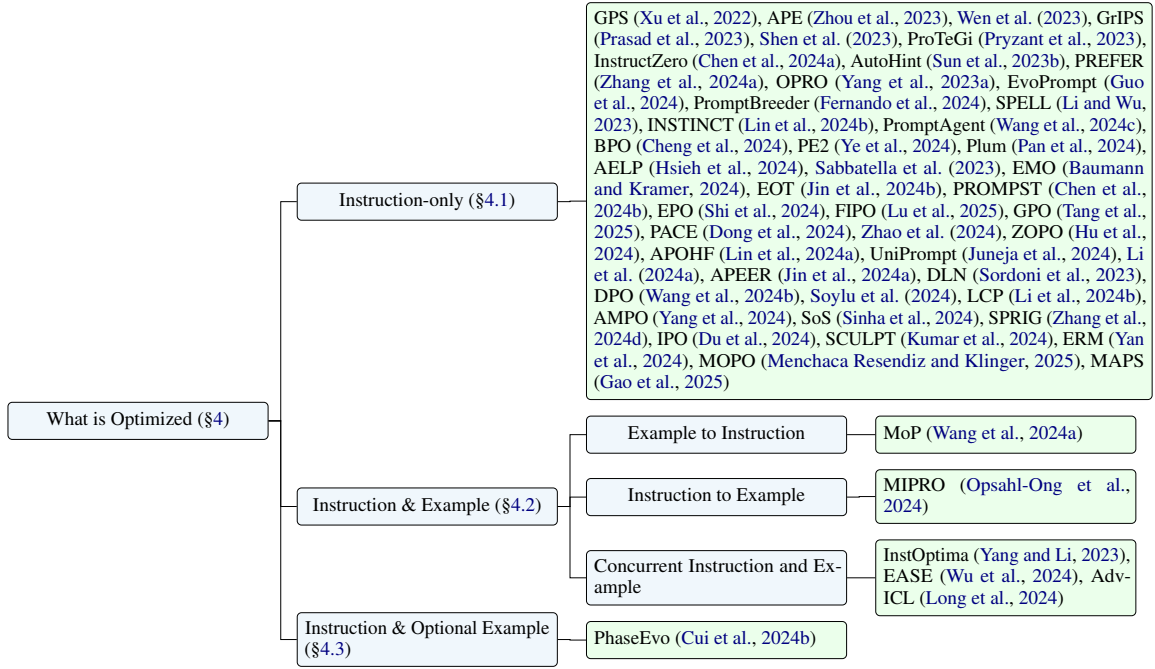
\begin{figure}[ht]
    \centering
    \resizebox{0.95\textwidth}{!}{
        \begin{forest}
            for tree={
                grow=east,
                draw,
                text width=40mm,
                font=\small,
                edge path={
                    \noexpand\path [draw, \forestoption{edge}] (!u.parent anchor) -- +(2mm,0) |- (.child anchor)\forestoption{edge label};
                },
                parent anchor=east,
                child anchor=west,
                tier/.wrap pgfmath arg={tier #1}{level()},
                edge={ultra thin},
                rounded corners=2pt,
                align=center,
                text badly centered,
                sibling distance=20mm
            }
            [What is Optimized (§\ref{sec:what_is_optimized}), highlight7,
                [\parbox{40mm}{Instruction \& Optional Example (§\ref{sec:instruction_optional_example})},
                    [\parbox{40mm}{PhaseEvo \cite{cui2024phaseevo}}, highlight1]
                ],
                [Instruction \& Example (§\ref{sec:instructandexample}),
                    [\parbox{40mm}{Concurrent Instruction and Example},
                        [\parbox{40mm}{
                            InstOptima \cite{yang2023instoptima}, EASE \cite{wu2024ease}, Adv-ICL \cite{advicl2024promptoptimizationadversarialincontext}
                        }, highlight1]
                    ],
                    [Instruction to Example,
                        [\parbox{40mm}{MIPRO \cite{mipro2024op}}, highlight1]
                    ],
                    [Example to Instruction,
                        [\parbox{40mm}{MoP \cite{wang2024mop}}, highlight1]
                    ]
                ],
                [Instruction-only (§\ref{sec:instructonly}),
                    [\parbox{90mm}{
                        GPS \cite{xu2022gps}, APE \cite{Zhou2023APE}, \citet{wen2023hardpromptseasygradientbased}, 
                        GrIPS \cite{prasad2022grips}, \citet{shen2023free},  
                        ProTeGi \cite{pryzant2023APO}, InstructZero \cite{Chen2023InstuructZero}, 
                        AutoHint \cite{sun2023autohint}, PREFER \cite{zhang2023peft}, OPRO \cite{Yang2023OPRO}, EvoPrompt \cite{Guo2023EVOPrompt}, PromptBreeder \cite{Fernando2023PromptBreeder}, SPELL \cite{li2023spell}, INSTINCT \cite{lin2024useinstinct}, PromptAgent \cite{wang2023promptagent}, BPO \cite{bpo2024cheng}, PE2 \cite{ye2024pe2}, Plum \cite{pan2024plum}, AELP \cite{Hsieh2023AELP}, \citet{sabbatella2023bayesian}, EMO \cite{baumann2024emo}, EOT \cite{jin2024eot}, PROMPST \cite{promptst2024chen}, EPO \cite{bandit2024shi}, FIPO \cite{lu2024fipo}, GPO \cite{tang2025gpo}, PACE \cite{dong-etal-2024-pace}, \citet{ps2024Zhao}, ZOPO \cite{huw2024zopo}, APOHF \cite{lin2024apohf}, UniPrompt \cite{juneja2024uniprompt}, \citet{concentrate2024li}, APEER \cite{jin2024apeer}, DLN \cite{dln2023sordoni}, DPO \cite{dpo2024want}, \citet{together2024soylu}, LCP \cite{li2024lcp}, AMPO \cite{yang2024ampo}, SoS \cite{sinha2024sos}, SPRIG \cite{zhang2024sprig}, IPO \cite{ipo2024du}, SCULPT \cite{kumar2024sculpt}, ERM \cite{yan2024erm}, MOPO \cite{mopo2024resendiz}, MAPS \cite{gao2025maps}
                    }, text width=90mm, tier=odd, highlight1]
                ]
            ]
        \end{forest}
    }
    \caption{{\textit{"What is optimized"} Categorization}}
    \label{fig:optimized}
\end{figure}

\begin{figure}[ht]
    \centering
    \resizebox{0.95\textwidth}{!}{
        \begin{forest}
            for tree={
                grow=east,
                draw,
                text width=40mm,
                font=\small,
                edge path={
                    \noexpand\path [draw, \forestoption{edge}] (!u.parent anchor) -- +(2mm,0) |- (.child anchor)\forestoption{edge label};
                },
                parent anchor=east,
                child anchor=west,
                tier/.wrap pgfmath arg={tier #1}{level()},
                edge={ultra thin},
                rounded corners=2pt,
                align=center,
                text badly centered,
                sibling distance=20mm
            }
            [What Criteria to Optimize (§\ref{sec:criteria_optimize}), highlight7,
                [Multi-Objective,
                    [\parbox{40mm}{InstOptima \cite{yang2023instoptima}, EMO \cite{baumann2024emo}, MOPO \cite{mopo2024resendiz}, SOS \cite{sinha2024sos}}, highlight1]
                ],
                [Safety and Ethical Constraints,
                    [\parbox{40mm}{GCG \cite{zou2023gcg}, RPO \cite{rpo2024zhou}}, highlight1]
                ],
                [Generalizability,
                    [\parbox{40mm}{\citet{concentrate2024li}}, highlight1]
                ],
                [Task Performance,
                    [\parbox{90mm}{
                        GPS \cite{xu2022gps}, APE \cite{Zhou2023APE}, \citet{wen2023hardpromptseasygradientbased}, 
                        GrIPS \cite{prasad2022grips}, \citet{shen2023free},  
                        ProTeGi \cite{pryzant2023APO}, InstructZero \cite{Chen2023InstuructZero}, 
                        AutoHint \cite{sun2023autohint}, PREFER \cite{zhang2023peft}, OPRO \cite{Yang2023OPRO}, EvoPrompt \cite{Guo2023EVOPrompt}, PromptBreeder \cite{Fernando2023PromptBreeder}, SPELL \cite{li2023spell}, INSTINCT \cite{lin2024useinstinct}, PromptAgent \cite{wang2023promptagent}, BPO \cite{bpo2024cheng}, PE2 \cite{ye2024pe2}, Plum \cite{pan2024plum}, AELP \cite{Hsieh2023AELP}, \citet{sabbatella2023bayesian}, EOT \cite{jin2024eot}, PROMPST \cite{promptst2024chen}, EPO \cite{bandit2024shi}, FIPO \cite{lu2024fipo}, GPO \cite{tang2025gpo}, PACE \cite{dong-etal-2024-pace}, \citet{ps2024Zhao}, ZOPO \cite{huw2024zopo}, APOHF \cite{lin2024apohf}, UniPrompt \cite{juneja2024uniprompt}, \citet{concentrate2024li}, APEER \cite{jin2024apeer}, DLN \cite{dln2023sordoni}, DPO \cite{dpo2024want}, \citet{together2024soylu}, LCP \cite{li2024lcp}, AMPO \cite{yang2024ampo}, SPRIG \cite{zhang2024sprig}, IPO \cite{ipo2024du}, SCULPT \cite{kumar2024sculpt}, ERM \cite{yan2024erm}, MAPS \cite{gao2025maps}
                        }, text width=90mm, tier=odd, highlight1]
                ]
            ]
        \end{forest}
    }
    \caption{{\textit{"What criteria to optimize"} Categorization}}
    \label{fig:objective}
\end{figure}
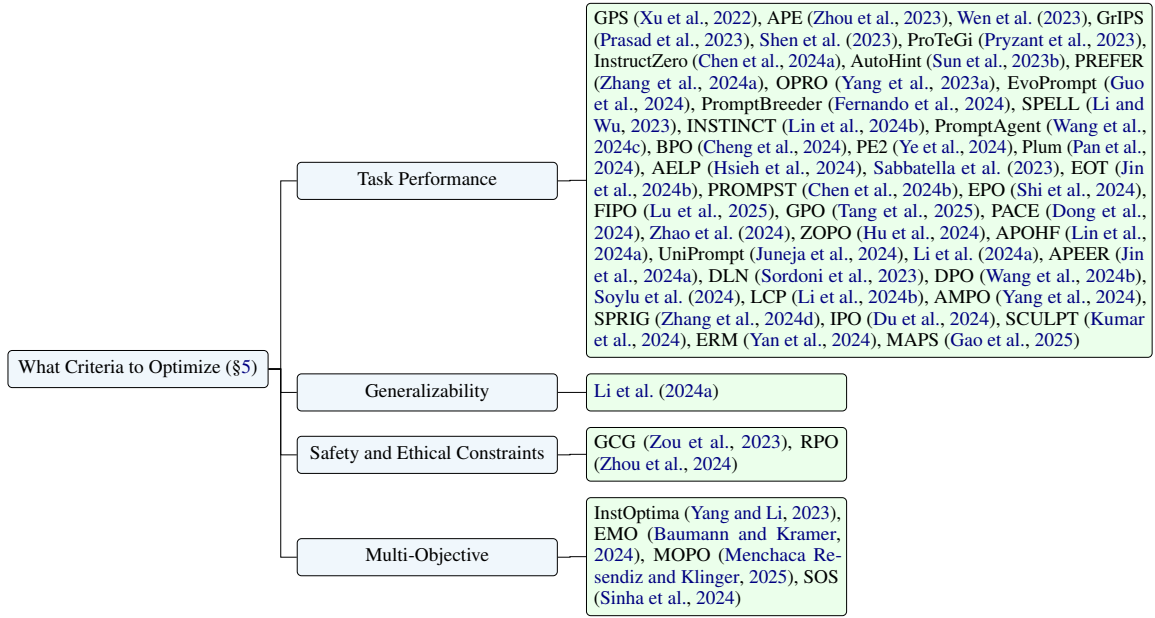

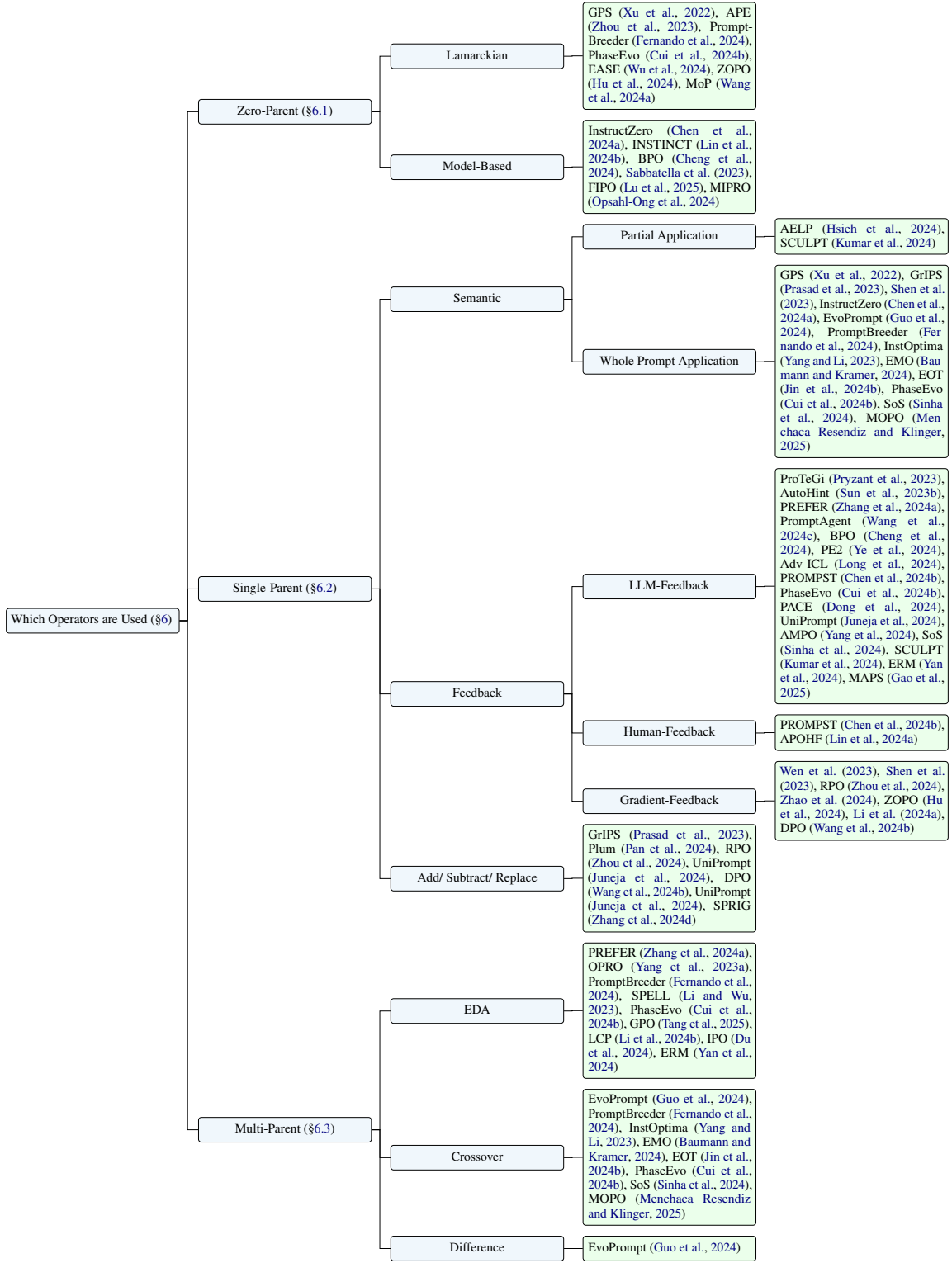
\begin{figure}[ht]
    \centering
    \resizebox{0.95\textwidth}{!}{ 
        \begin{forest}
            for tree={
                grow=east,
                draw,
                text width = 40mm,
                font=\small,
                edge path={
                    \noexpand\path [draw, \forestoption{edge}] (!u.parent anchor) -- +(2mm,0) |- (.child anchor)\forestoption{edge label};
                },
                parent anchor=east,
                child anchor=west,
                tier/.wrap pgfmath arg={tier #1}{level()},
                edge={ultra thin},
                rounded corners=2pt,
                align=center,
                text badly centered, 
                sibling distance=20mm
            }
            [\parbox{40mm}{Which Operators are Used (§\ref{sec:operators})}, highlight7,
                [Multi-Parent (§\ref{sec:multiparent}),
                    [Difference,
                        [\parbox{40mm}{EvoPrompt \cite{Guo2023EVOPrompt}}, highlight1]    
                    ]
                    [Crossover,
                       [\parbox{40mm}{EvoPrompt \cite{Guo2023EVOPrompt}, PromptBreeder \cite{Fernando2023PromptBreeder}, InstOptima \cite{yang2023instoptima}, EMO \cite{baumann2024emo}, EOT \cite{jin2024eot}, PhaseEvo \cite{cui2024phaseevo}, SoS \cite{sinha2024sos}, MOPO \cite{mopo2024resendiz}}, highlight1]
                    ]
                    [EDA,
                        [\parbox{40mm}{PREFER \cite{zhang2023peft}, OPRO \cite{Yang2023OPRO}, PromptBreeder \cite{Fernando2023PromptBreeder}, SPELL \cite{li2023spell}, PhaseEvo \cite{cui2024phaseevo}, GPO \cite{tang2025gpo}, LCP \cite{li2024lcp}, IPO \cite{ipo2024du}, ERM \cite{yan2024erm}}, highlight1]
                    ]
                ]
                [Single-Parent (§\ref{sec:singleparent}),
                    [Add/ Subtract/ Replace,
                        [\parbox{40mm}{GrIPS \cite{prasad2022grips}, Plum \cite{pan2024plum}, RPO \cite{rpo2024zhou}, UniPrompt \cite{juneja2024uniprompt}, DPO \cite{dpo2024want},  UniPrompt \cite{juneja2024uniprompt}, SPRIG \cite{zhang2024sprig}}, highlight1]
                    ]
                    [Feedback,
                        [Gradient-Feedback,
                            [\parbox{40mm}{\citet{wen2023hardpromptseasygradientbased}, \citet{shen2023free}, RPO \cite{rpo2024zhou}, \citet{ps2024Zhao}, ZOPO \cite{huw2024zopo}, \citet{concentrate2024li}, DPO \cite{dpo2024want}}, highlight1]
                        ]
                        [Human-Feedback,
                            [\parbox{40mm}{PROMPST \cite{promptst2024chen}, APOHF \cite{lin2024apohf}}, highlight1]
                        ]
                        [LLM-Feedback,
                            [\parbox{40mm}{ProTeGi \cite{pryzant2023APO}, AutoHint \cite{sun2023autohint}, PREFER \cite{zhang2023peft}, PromptAgent \cite{wang2023promptagent}, BPO \cite{bpo2024cheng}, PE2 \cite{ye2024pe2}, Adv-ICL \cite{advicl2024promptoptimizationadversarialincontext}, PROMPST \cite{promptst2024chen}, PhaseEvo \cite{cui2024phaseevo}, PACE \cite{dong-etal-2024-pace}, UniPrompt \cite{juneja2024uniprompt}, AMPO \cite{yang2024ampo}, SoS \cite{sinha2024sos}, SCULPT \cite{kumar2024sculpt}, ERM \cite{yan2024erm}, MAPS \cite{gao2025maps}}, highlight1]
                        ]
                    ]
                    [Semantic,
                        [Whole Prompt Application,
                            [\parbox{40mm}{GPS \cite{xu2022gps}, GrIPS \cite{prasad2022grips}, \citet{shen2023free}, InstructZero \cite{Chen2023InstuructZero}, EvoPrompt \cite{Guo2023EVOPrompt}, PromptBreeder \cite{Fernando2023PromptBreeder}, InstOptima \cite{yang2023instoptima}, EMO \cite{baumann2024emo}, EOT \cite{jin2024eot}, PhaseEvo \cite{cui2024phaseevo}, SoS \cite{sinha2024sos}, MOPO \cite{mopo2024resendiz}}, highlight1]
                        ]
                        [Partial Application,
                             [\parbox{40mm}{ AELP \cite{Hsieh2023AELP}, SCULPT \cite{kumar2024sculpt}}, highlight1]
                        ]
                    ]
                ]
                [Zero-Parent (§\ref{sec:zeroparent}),
                    [Model-Based,
                        [\parbox{40mm}{InstructZero \cite{Chen2023InstuructZero}, INSTINCT \cite{lin2024useinstinct}, 
                        BPO \cite{bpo2024cheng}, \citet{sabbatella2023bayesian}, FIPO \cite{lu2024fipo}, MIPRO \cite{mipro2024op}}, highlight1]
                    ]
                    [Lamarckian,
                        [\parbox{40mm}{GPS \cite{xu2022gps}, APE \cite{Zhou2023APE}, PromptBreeder \cite{Fernando2023PromptBreeder}, PhaseEvo \cite{cui2024phaseevo}, EASE \cite{wu2024ease}, ZOPO \cite{huw2024zopo}, MoP \cite{wang2024mop}}, highlight1]
                    ]
                ]
            ]
        \end{forest}
    }
    \caption{{\textit{"Which operators are used"} Categorization }}
    \label{fig:operator}
\end{figure}

\begin{figure}[ht]
    \centering
    \resizebox{0.95\textwidth}{!}{ 
        \begin{forest}
            for tree={
                grow=east,
                draw,
                text width=40mm,
                font=\small,
                edge path={
                    \noexpand\path [draw, \forestoption{edge}] (!u.parent anchor) -- +(2mm,0) |- (.child anchor)\forestoption{edge label};
                },
                parent anchor=east,
                child anchor=west,
                tier/.wrap pgfmath arg={tier #1}{level()},
                edge={ultra thin},
                rounded corners=2pt,
                align=center,
                text badly centered,
                sibling distance=20mm
            }
            [\parbox{40mm}{Which Iterative Algorithm is Used (§\ref{sec:iteration_algorithms})}, highlight7,
                [Iterative Refinement (§\ref{sec:iterative}),
                    [\parbox{40mm}{\citet{wen2023hardpromptseasygradientbased}, \citet{shen2023free}, BPO \cite{bpo2024cheng}, \citet{sabbatella2023bayesian},
                    Adv-ICL \cite{advicl2024promptoptimizationadversarialincontext}, FIPO \cite{lu2024fipo}, ZOPO \cite{huw2024zopo}, \citet{concentrate2024li}, DPO \cite{dpo2024want}}, highlight1]
                ],
                [Metaheuristic Algorithm (§\ref{sec:metahurisitc}),
                    [Phased Algorithm,
                        [\parbox{40mm}{PhaseEvo \cite{cui2024phaseevo}}, highlight1]
                    ],
                    [General Metaheuristic,
                        [\parbox{40mm}{InstructZero \cite{Chen2023InstuructZero}, AutoHint \cite{sun2023autohint}, PREDER \cite{zhang2023peft}, OPRO \cite{Yang2023OPRO}, PE2 \cite{ye2024pe2}, Plum \cite{pan2024plum}, RPO \cite{rpo2024zhou}, GPO \cite{tang2025gpo}, \citet{ps2024Zhao}, Uniprompt \cite{juneja2024uniprompt}, MIPRO \cite{mipro2024op}, APEER \cite{jin2024apeer}, DLN \cite{dln2023sordoni}, LCP \cite{li2024lcp}, AMPO \cite{yang2024ampo}, SoS \cite{sinha2024sos}, IPO \cite{ipo2024du}, MAPS \cite{gao2025maps}}, highlight1]
                    ],
                    [Evolutionary Algorithm,
                        [Differential Evolution,
                            [\parbox{40mm}{EvoPrompt \cite{Guo2023EVOPrompt}}, highlight1]
                        ],
                        [Genetic Algorithm,
                            [\parbox{40mm}{GPS \cite{xu2022gps}, \citet{shen2023free}, EvoPrompt \cite{Guo2023EVOPrompt}, PromptBreeder \cite{Fernando2023PromptBreeder}, SPELL \cite{li2023spell}, INSTINCT \cite{lin2024useinstinct}, InstOptima \cite{yang2023instoptima}, EMO \cite{baumann2024emo}, EOT \cite{jin2024eot}, PACE \cite{dong-etal-2024-pace}, DPO \cite{dpo2024want}, SPRIG \cite{zhang2024sprig}, MOPO \cite{mopo2024resendiz}}, highlight1]
                        ]
                    ]
                ],
                [Monte Carlo Search (§\ref{sec:montecarlo}),
                    [Monte Carlo Tree Search,
                        [\parbox{40mm}{PromptAgent \cite{wang2023promptagent}}, highlight1]
                    ],
                    [Monte Carlo Search,
                        [\parbox{40mm}{APE \cite{Zhou2023APE}}, highlight1]
                    ]
                ],
                [Heuristic Sampling (§\ref{sec:heuristicsampling}),
                    [\parbox{40mm}{INSTINCT \cite{lin2024useinstinct}, PROMPST \cite{promptst2024chen}, MoP \cite{wang2024mop}}, highlight1]
                ],
                [Beam Search (§\ref{sec:beam}),
                    [\parbox{40mm}{GrIPS \cite{prasad2022grips}, ProTeGi \cite{pryzant2023APO}, AELP \cite{Hsieh2023AELP}, SCULPT \cite{kumar2024sculpt}, ERM \cite{yan2024erm}}, highlight1]
                ],
                [Bandit Algorithm (§\ref{sec:bandit}),
                    [\parbox{40mm}{EPO \cite{bandit2024shi}, EASE \cite{wu2024ease}, APOHF \cite{lin2024apohf}}, highlight1]
                ]
            ]
        \end{forest}
    }
    \caption{{\textit{"Which iterative algorithm is used"} Categorization }}
    \label{fig:algoithm}
\end{figure}
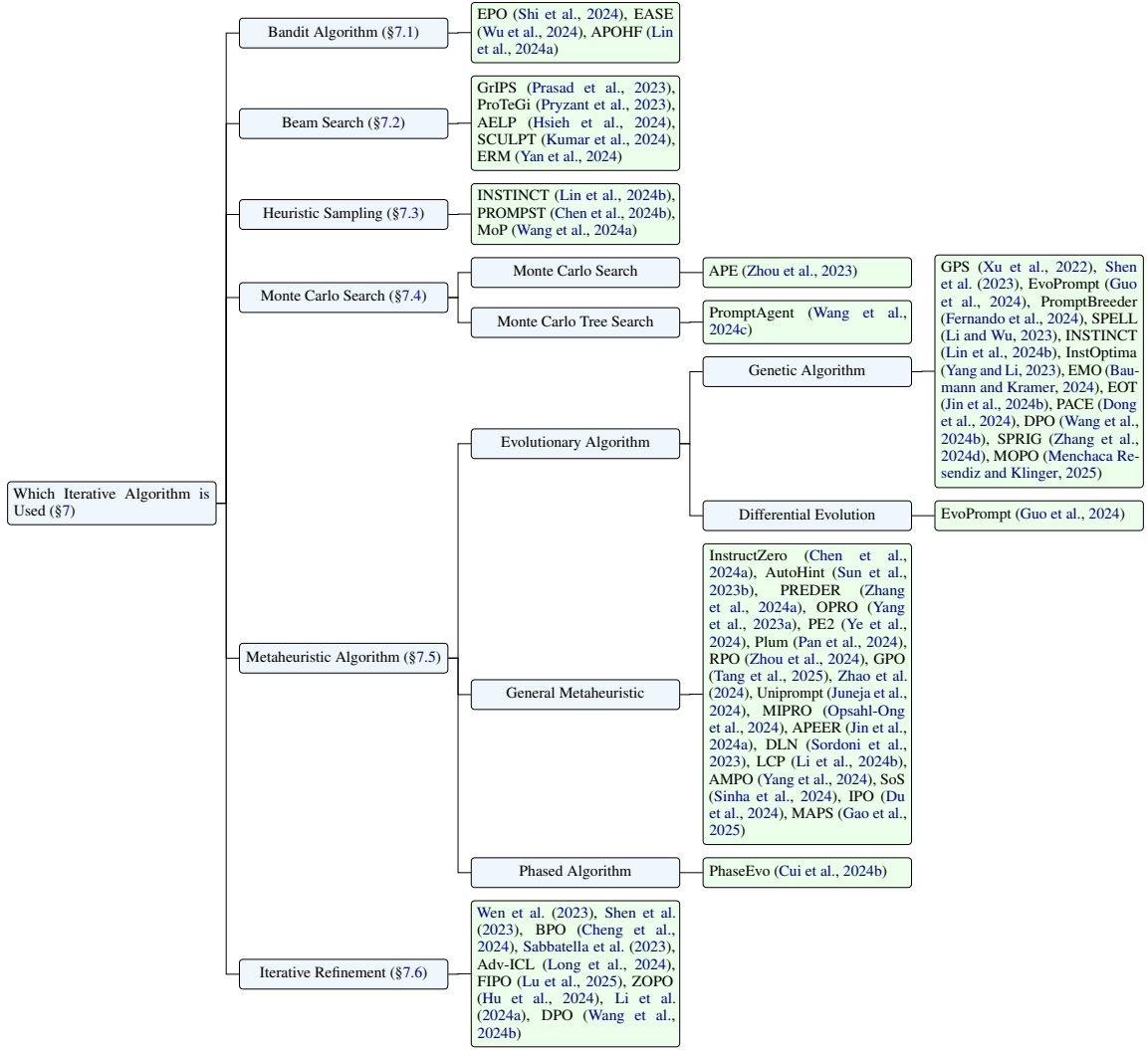

\clearpage

\section {Datasets and Tools}
\label{sec:datasets_tools}

\begin{table}[ht]
\small
    \centering
    \begin{tabular}{@{}l|l@{}}
        \toprule
        \textbf{Dataset Name} & \textbf{Dataset Category}  \\
        \midrule
        BBH \citep{srivastava2023beyond} & NLP Benchmark \\
        Instruction Induction \cite{Honovich2022Instuct} & NLP Benchmark \\
        GSM8K \cite{cobbe2021gsm8k} & Mathematical Reasoning \\
        Ethos \cite{mollas2022ethos} & Bias and Ethics Evaluation  \\
        SST-2 \cite{socher2013sst2} & Sentiment Analysis \\
        HotpotQA \cite{yang2018hotpotqa} & Question Answering  \\
        Iris \cite{li2024iris} & Scientific Classification \\
        SVAMP \cite{patel2021svamp} & Mathematical Reasoning\\
        Subj \cite{pang2004subj} & Subjectivity Detection  \\
        CR \cite{hu2004cr} & Sentiment Analysis  \\
        MR \cite{pang2005mr} & Sentiment Analysis \\
        TREC \cite{voorhees2000trec} & Question Answering  \\
        Liar \cite{wang2017liar} & Misinformation Detection  \\
        \bottomrule
    \end{tabular}
    \caption{Commonly used datasets}
    \label{tab:dataset_categorization}
\end{table}

\end{document}